\documentclass[letterpaper]{article} 
\usepackage[preprint]{aaai2027}  
\usepackage[hyphens]{url}  
\usepackage{graphicx} 
\usepackage{natbib}  
\usepackage{caption} 
\usepackage{algorithm}
\usepackage{algorithmic}
\usepackage{amsmath, amssymb, amsthm}
\usepackage{multirow}
\newtheorem{definition}{Definition}
\newtheorem{example}{Example}
\newtheorem{problem}{Problem}
\usepackage{newfloat}
\usepackage{listings}
\DeclareCaptionStyle{ruled}{labelfont=normalfont,labelsep=colon,strut=off} 
\floatstyle{ruled}
\newfloat{listing}{tb}{lst}{}
\floatname{listing}{Listing}

\usepackage{booktabs}

\title{MiGHT-EHR: A Multi-task Graph Transformer for Heterogeneous Temporal Electronic Health Records}
\author{
    Anirudh Rayas\textsuperscript{\rm 1},
    Yuan Wang\textsuperscript{\rm 2},
    Pavan Turaga\textsuperscript{\rm 1}
}
\affiliations{
   \textsuperscript{\rm 1}Arizona State University, Tempe, AZ, USA\\
    \textsuperscript{\rm 2}University of South Carolina, Columbia, SC, USA\\
    \{ahrayas, pturaga\}@asu.edu,
    wang578@mailbox.sc.edu
}

\begin{document}

\maketitle

\begin{abstract}
 Learning from Electronic Health Records (EHRs) has gained significant attention due to its potential to improve clinical prediction. However, effective learning remains challenging because EHRs encode heterogeneous, temporally ordered clinical interactions. In particular, EHRs contain: (i) heterogeneous clinical entities, including patients, visits, diagnoses, prescriptions, and procedures, together with their heterogeneous interactions, (ii) longitudinal patient trajectories across hospital visits and (iii) shared statistical dependencies across related clinical prediction tasks. Existing EHR learning methods capture only a subset of these properties. To bridge this gap, we propose Multi-task Graph transformer for Heterogeneous Temporal EHRs (MiGHT-EHR), which jointly models all three within a unified representation learning method. MiGHT-EHR constructs a heterogeneous graph from EHRs in which nodes represent clinical entities and edges connect statistically associated entities identified via normalized point-wise mutual information. Across MIMIC-III and MIMIC-IV datasets, MiGHT-EHR outperforms state-of-the-art methods on average across four tasks: drug recommendation, prediction of length-of-stay, mortality, and readmission, with particularly strong improvements in mortality and readmission prediction. Furthermore, a post-hoc analysis of the learned representations reveals that patient neighborhoods are organized by clinical outcomes, salient medical concepts are recoverable as linear directions in the representation space, and task probabilities are well calibrated. Collectively, these findings demonstrate that MiGHT-EHR representations support diverse prediction tasks while preserving clinically interpretable structure.
\end{abstract}


\section{Introduction}
\label{sec:intro}

Electronic health records (EHRs) are not merely collections of isolated clinical observations of a patient, but rather encode rich relationships among clinical entities present in patient records. Learning effective representations from such data has received significant attention as it enables diverse clinical decision-support tasks~\cite{xiao2018opportunities, rajkomar2018scalable, shickel2017deep, cheng2016risk}. As EHRs continue to grow in scale and complexity, the challenge is learning clinical context from underlying graphical structure that connects these observations.

This structure arises from three complementary sources. First, EHRs contain heterogeneous clinical entities like patients, hospital visits, diagnoses, procedures, prescriptions, and laboratory tests, each connected through distinct clinical relationships. Second, these entities evolve over time as patients accumulate longitudinal healthcare trajectories spanning multiple hospital visits. Finally, many clinical prediction tasks are intrinsically related because they are learned from the same patient population and therefore share statistical dependencies that can be exploited through joint learning. Together, these structural properties provide rich contextual information for representation learning.

Existing methods capture only subsets of this structure. Sequence-based models capture temporal features in EHR data but overlook the relationships among heterogeneous clinical entities~\cite{choi2016doctor, ma2017dipole, rasmy2021med}. Conversely, heterogeneous graph-based models capture relationships among clinical entities but often ignore temporal patient trajectories or are designed for a single prediction task~\cite{choi2020learning, liu2020hsgnn, jung2024graphehr}. Jointly modeling heterogeneous clinical entities, longitudinal patient trajectories, and shared dependencies across multiple clinical prediction makes learning from EHR substantially more challenging.

To address this challenge, we propose \textbf{MiGHT-EHR} (Multi-task Graph Transformer for Heterogeneous Temporal Electronic Health Records), a novel approach for multi-task learning from heterogeneous temporal EHRs. MiGHT-EHR first constructs a heterogeneous temporal EHR graph based on statistical associations among clinical entities extracted from patient records. A heterogeneous graph transformer with temporal attention then learns shared representations for multi-task learning. Our main contributions are summarized as follows.

\begin{itemize}
\item We introduce a graph construction methodology that transforms patient records into a heterogeneous temporal EHR graph (Def.~\ref{def:httg}), where clinical entities are connected through three complementary relation types (Sec.~\ref{sec:construction}): membership relations that capture the EHR structure, temporal relations that encode longitudinal patient trajectories, and statistically associated co-occurrence relations among clinical entities.
\item We design a self-supervised pretraining module (Sec.~\ref{sec:features}) that enhances the nodes' relational features in the EHR graph. We then use a heterogeneous graph transformer (Sec.~\ref{sec:mp}) equipped with temporal attention to learn node-level features .
\item We employ dual-balancing multi-task learning (DB-MTL) framework in our training procedure to equalize each task's gradient contribution so that all tasks are learned equally.
\item We demonstrate the effectiveness of MiGHT-EHR on MIMIC-III and MIMIC-IV benchmark datasets (Sec.~\ref{sec:experiments}), where it outperforms state-of-the-art methods on average across four tasks: drug recommendation and the prediction of length of stay, mortality, and readmission. A post-hoc analysis of the learned representations shows that patient neighborhoods are organized by clinical outcome, salient medical concepts are recoverable as linear directions, and predicted probabilities are well calibrated.
\end{itemize}

\section{Related Work}
\paragraph{Graph Representation Learning for EHRs:} 
EHRs encode rich relational information between medical entities (patients make visits, visits record diagnoses, prescriptions, and procedures) and have become a valuable resource for deep learning-based clinical prediction~\cite{xiao2018opportunities, rajkomar2018scalable, shickel2017deep, cheng2016risk}. Increasingly graph-based models such as graph neural networks (GNNs)~\cite{boll2024graph, paul2024systematic, choi2020learning} have been used to exploit the inherent relational structure. Early methods modeled EHRs as homogeneous graphs~\cite{choi2017gram, yao2024self, wu2021EERM}, but different medical entities play distinct roles and are connected by different semantic relations, motivating heterogeneous GNN formulations~\cite{liu2020hsgnn, wanyan2020heterogeneous, jung2024graphehr, wang2024faircare} (see Def.~\ref{def:hetgraph}). These, however, ignore the temporal dependency between a patient's consecutive visits. Sequence based models capture temporal order but discards the relationships among clinical entities~\cite{ma2017dipole, nguyen2016deepr, ma2020concare, gao2020stagenet}. We instead propose a heterogeneous graph transformer model with temporal attention to jointly capture heterogeneous entity interactions and temporal dependencies in a unified representation.

\paragraph{Graph Construction and Multi-task Learning:}
The preceding methods assume the graph is given, yet its edges must first be constructed. Knowledge graphs built from EHRs 
typically weight edges by co-occurrence frequency~\cite{mythili2022construction, murali2023towards, li2020real}, which is confounded by entity frequency since common entities co-occur frequently, so edges reflect popularity rather than association. We instead form edges via normalized point-wise mutual information (NPMI, eq.~\eqref{eq:npmi}), linking two entities only when NPMI values are greater than a certain threshold. Given such a graph, a shared encoder can serve several clinical tasks by exploiting their relational structure. Prior multi-task learning methods~\cite{chan_multitask_ehr, hur2023genhpf, cui2024automated}, treats the task losses uniformly. In practice, however, tasks differ in loss scale and gradient magnitude, so the largest-gradient task dominates the shared encoder. We adopt Dual-balancing for multi-task learning (DB-MTL) ~\cite{lin2023dbmtl}, which equalizes per-task gradient scale and magnitude, so that every task contributes equally.

\section{Preliminaries and Problem Formulation}
\label{sec:prelim}

We begin by introducing heterogeneous graphs and extend this notion to heterogeneous temporal EHR graphs. We then formulate multi-task learning over such graphs and conclude with the problem studied in this paper.

\begin{definition}[Heterogeneous graph]
\label{def:hetgraph}
A \emph{heterogeneous graph} is a tuple
$\mathcal{G} = (\mathcal{V}, \mathcal{E}, \phi, \psi)$ with node set
$\mathcal{V}$, edge set $\mathcal{E} \subseteq \mathcal{V} \times \mathcal{V}$,
a node-type map $\phi : \mathcal{V} \to \mathcal{T}_V$, and a relation-type map
$\psi : \mathcal{E} \to \mathcal{T}_E$. The graph $\mathcal{G}$ is heterogeneous when $|\mathcal{T}_V| + |\mathcal{T}_E| > 2$.
\end{definition}


While Definition~\ref{def:hetgraph} captures the heterogeneous structure of EHR data, it does not account for its temporal nature. In practice, each patient accrues a chronologically ordered sequence of visits, and this temporal ordering carries predictive information. The following definition extends the heterogeneous graph formalism to heterogeneous temporal EHR graphs.

\begin{definition}[Heterogeneous Temporal EHR Graph]
\label{def:httg}
A \emph{heterogeneous temporal EHR graph} is a heterogeneous graph
$\mathcal{G}=(\mathcal{V},\mathcal{E},\phi,\psi)$ (Def.~\ref{def:hetgraph}) where each visit node $v\in\mathcal{T}_{V}$ is annotated with a
timestamp $t_v\in[0,1]$, and a temporal relation
$\texttt{next\_visit}\in\mathcal{T}_E$ that links the temporally consecutive visits of
each patient. More precisely, for two visits $u,v$ of the same patient,
\begin{align*}
\psi(u,v)=&\texttt{next\_visit}
\;\iff\;
t_u \le t_v \ \text{ and }\ \\ & \nexists\, w :\ t_u < t_w < t_v,
\end{align*}
where $w$ ranges over that patient's visits.
\end{definition}

 The following example illustrates how heterogeneous entity relations and temporal visit dependencies coexist within a heterogeneous temporal EHR graph.

\begin{example}
Consider a visit $v$ that is diagnosed ($d$) with
type~2 diabetes and metformin as prescription ($m$). This induces the edges $\psi(v,d)=\texttt{diagnosed}$ and
$\psi(v,m)=\texttt{prescribed}$. Consequently, the neighborhood $\mathcal{N}_{\texttt{diagnosed}}(d)$ consists of all
visits diagnosed with diabetes. If the same patient is readmitted at a later visit $v'$ with $t_v<t_{v'}$, a temporal edge
$\psi(v,v')=\texttt{next\_visit}$ connects the two visits, allowing the representation of $v'$ to incorporate information from the earlier visit $v$.
\end{example}

Having defined the heterogeneous temporal EHR graph, we now describe the learning setting considered throughout the paper. Specifically, we formulate clinical prediction as a multi-task learning problem over visit representations.
\paragraph{Multi-task learning over visits.}
We adopt a multi-task formulation in which a
single encoder produces one representation per visit, shared across all tasks.
This suits learning on
EHRs because the tasks are not independent: they are distinct
consequences of the same underlying clinical state, so a representation
informative for one task tends to be informative for another, and a shared
encoder lets the tasks regularize one another. The tasks are nonetheless
heterogeneous, spanning binary, ordinal, and multi-label targets on different
scales, so their losses must be combined with care rather than simply summed. We formalize this setting as the following optimization problem.

\begin{problem}[Multi-task learning on a heterogeneous temporal EHR graph]
\label{prob:mtl} 
Given a heterogeneous temporal EHR graph $\mathcal{G}$ (Def.~\ref{def:httg}) and
$K$ tasks, where tasks $k = \{1,\cdots,K\}$
is specified by a label map
$y^k : \mathcal{V}^k \to \mathcal{Y}_k$ over its labeled visits
$\mathcal{V}^k \subseteq \mathcal{V}_{\mathrm{vis}}$. The goal is to learn a single shared
encoder $f_\theta : (\mathcal{G}, \mathcal{V}_{\mathrm{vis}}) \to \mathbb{R}^{d}$
and $K$ task-specific heads $g_{\varphi_k} : \mathbb{R}^{d} \to \mathcal{Y}_k$
that minimize
\begin{align}
&\min_{\theta, \{\varphi_k\}} \;
\Phi\!\bigl( \mathcal{L}_1, \ldots, \mathcal{L}_K; \theta; \{\varphi_k\}\bigr) \quad \text{where},\\
&\mathcal{L}_k = \frac{1}{|\mathcal{V}^k|}
\sum_{v \in \mathcal{V}^k}
\ell_k\!\bigl( g_{\varphi_k}( f_\theta(\mathcal{G},v),\, y_k(v) \bigr),
\label{eq:mtl}
\end{align}
where $\theta$ is shared across all tasks, $\ell_k$ is the per-task loss, and
$\Phi$ combines the task losses.
\end{problem}

This formulation provides the foundation for the proposed Multi-task Graph transformer for Heterogeneous Temporal EHR (MiGHT-EHR) method, which specifies the construction of the heterogeneous temporal graph, the shared encoder, and the task-balancing optimization strategy.

\section{Methodology}
\label{sec:method}

\begin{figure*}[t]
  \centering
  \includegraphics[width=\textwidth]{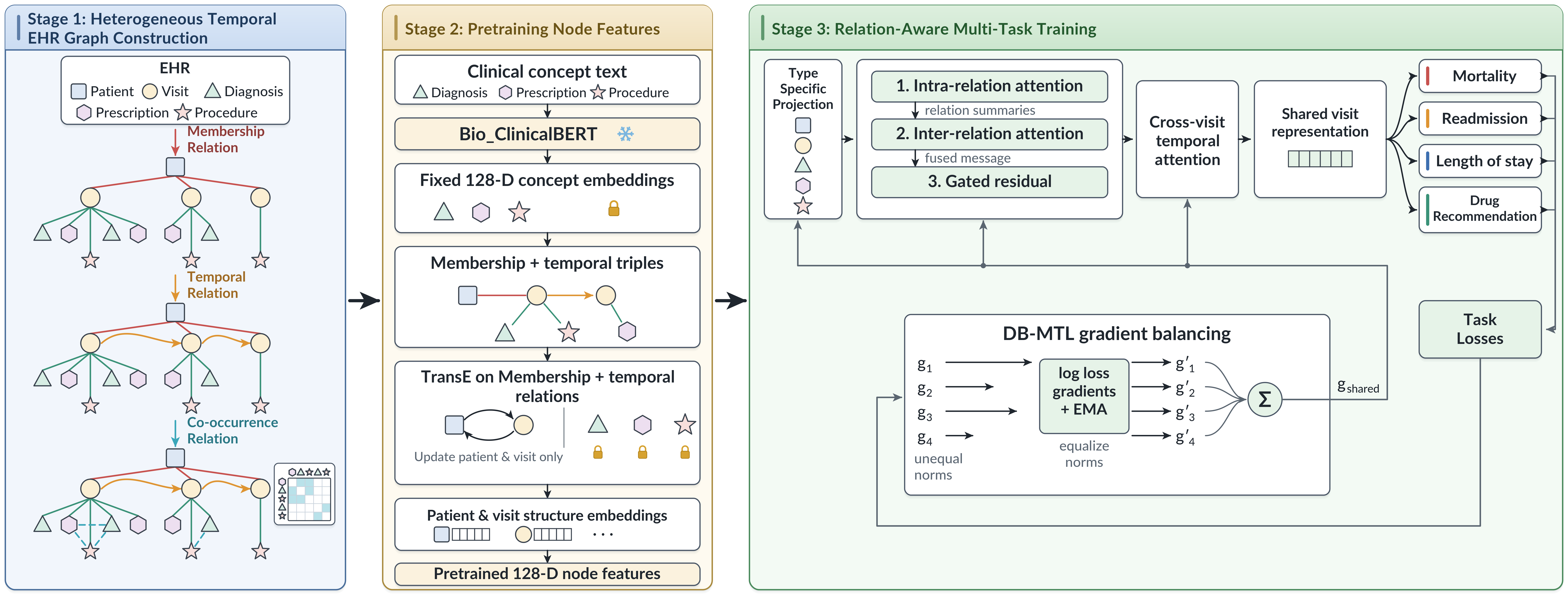}
  \caption{Overview of MiGHT-EHR. \textbf{Stage~1:} patient records become a
  graph of five node types, linked by \emph{membership} edges (edges linking each visit to its patient and its concept nodes), \emph{temporal} edges (directed edges between a patient's consecutive visits), and \emph{co-occurrence} edges (edges between co-occurring concept nodes determined by NPMI values). \textbf{Stage~2:} node features are pretrained and frozen:
  text embeddings for concepts, structural embeddings for patients and visits.
  \textbf{Stage~3:} a heterogeneous graph transformer with temporal attention
  builds one shared visit representation for all four tasks, trained with DB-MTL
  so that no single task dominates.}
  \label{fig:architecture}
\end{figure*}

In this section, we present the proposed method (MiGHT-EHR) 
in three stages, (i) constructing a heterogeneous temporal EHR graph from a clinical database (Section~\ref{sec:construction}), (ii) initializing its nodes with pretrained clinical features (Section~\ref{sec:features}), and (iii) encoding the graph with relation-aware message passing trained jointly across tasks (Section~\ref{sec:mp}).

\subsection{Heterogeneous Temporal EHR Graph} 
\label{sec:construction}

We instantiate the heterogeneous temporal EHR graph of Def.~\ref{def:httg} with five node types: \texttt{patient}, \texttt{visit}, \texttt{diagnosis}, \texttt{procedure}, and \texttt{prescription}. We refer to \texttt{diagnosis}, \texttt{procedure}, and \texttt{prescription} nodes collectively as \emph{concept nodes}. The graph connects the five node-types through three categories of relations: \emph{membership} (edges linking each visit to its patient and its concept nodes), \emph{temporal} (directed edges between a patient's consecutive visits), and \emph{co-occurrence} (edges between co-occurring concept nodes).

\paragraph{Membership relations.}
These edges denote connections between entities and visits.
Each visit node is connected to the patient node who generated it and 
the diagnoses, procedures, and prescriptions recorded during it, through four relations: \texttt{makes} (\texttt{patient}$\to$\texttt{visit}), \texttt{diagnosed}
(\texttt{visit}$\to$\texttt{diagnosis}), \texttt{treated}
(\texttt{visit}$\to$\texttt{procedure}), and \texttt{prescribed}
(\texttt{visit}$\to$\texttt{prescription}), each with an explicit reverse relation so that messages propagate in both directions during training.

\paragraph{Temporal relations.}
Each visit $i$ carries an admission time normalized to $t_i\in[0,1]$
over the observation window. For each
patient, visits are sorted by admission time and consecutive visits are linked
by the directed relation $\texttt{next\_visit}$ of Def.~\ref{def:httg}, left
without a reverse so that a visit is informed by its predecessors and not its
successors.

\paragraph{Co-occurrence edges.}
These edges connect concepts nodes that are clinically associated within a visit.
A naive rule connects two concepts co-occurring in at least $\kappa$
visits, but the raw count grows with how frequent each concept is individually:
common concepts co-occur frequently, so a count threshold retains
uninformative edges among frequent concepts while discarding rare but specific
pairings. We instead retain an edge only where two concepts co-occur
\emph{more than their individual frequencies predict},
measured by normalized
point-wise mutual information (NPMI),
\begin{equation}
\mathrm{NPMI}(a,b) \!=\! \frac{1}{-\log p(a,b)}\,\log \frac{p(a,b)}{p(a)\,p(b)} \;\in\; [-1,1],
\label{eq:npmi}
\end{equation}
where $p(a)=c(a)/N$ is the fraction of the $N=|\mathcal{V}_{\mathrm{vis}}|$
visits containing concept $a$ and $p(a,b)$ the fraction containing both. Thus, when $\mathrm{NPMI}=1$ (i.e., $p(a)=p(b)=p(a,b)$) the two concepts always co-occur, $0$ at independence, and negative when they are mutually exclusive. We retain a co-occurrence edge iff, $\mathrm{NPMI}(a,b)\geq\tau$ and $c(a,b)\geq\kappa$, we call this the co-occurrence criterion. The parameter $\tau$ removes the frequency confound and the count floor $\kappa$ removes small-sample noise, since the estimate from a handful of visits is unreliable
even though NPMI itself is scale-free. 

Applied within each concept type and across each pair of types,
the co-occurrence criterion yields nine co-occurrence relations (three same-type and three cross-type with reverses, see table in Appendix for complete graph statistics). Together with the eight
membership relations and $\texttt{next\_visit}$, the EHR graph carries
$|\mathcal{T}_E|=18$ relation types.

\subsection{Pretrained Node Features}
\label{sec:features}

Every node $i$ is assigned a feature vector
$\mathbf{x}_i\in\mathbb{R}^{d_{\mathrm{in}}}$ ($d_{\mathrm{in}}=128$), obtained by pretraining and held \emph{fixed} during the multi-task training of
Section~\ref{sec:mp}. The concept nodes are initialized with Bio\_ClinicalBERT~\cite{alsentzer2019publicly} encoder and the patient and visit nodes are initialized with TransE \cite{bordes2013transe} pretraining. Through membership relations, TransE aligns patient and visit features with their connected concept nodes thereby transferring clinical context to patients and visits. These pretrained features are held fixed during multi-task training in Section~\ref{sec:mp}.

\paragraph{Clinical concepts.}
Node types like \texttt{diagnosis}, \texttt{procedure}, and \texttt{prescription} are represented by the natural-language descriptions of the corresponding clinical concepts. Each description is encoded with a frozen
Bio\_ClinicalBERT~\cite{alsentzer2019publicly}, where token embeddings are
mean-pooled into a $768$-dimensional vector and reduced to $128$ dimensions by
principal component analysis. Encoding a concept's text rather than an arbitrary
identifier places clinically related concepts close in feature space and gives
even a rarely observed concept an informative representation.

\paragraph{Patients and visits.}
Patient and visit nodes carry no descriptive text, so we learn their features from the graph's structure using TransE~\cite{bordes2013transe} pretraining. The idea is to place each node in a vector space and treat every relation as a fixed shift, so that a true edge $(i,r,j)$ satisfies: $\mathbf{x}_i + \mathbf{r}_r \approx \mathbf{x}_j$. How well an edge fits this is measured by the distance $s(i,r,j) = \lVert \mathbf{x}_i + \mathbf{r}_r - \mathbf{x}_j \rVert$, which should be small for true edges. We train the embeddings so that each true edge scores a smaller distance, by a margin, than a \emph{corrupted} edge, obtained by swapping one of its endpoints for a random node. More precisely,
\begin{equation}
\sum_{(u,r,w)} \sum_{(u',r,w')}
\bigl[\, s(u,r,w) - s(u',r,w') + \gamma \,\bigr]_{+},
\end{equation}
where $(u',r,w')$ is a corrupted edge, 
$\gamma$ the margin, and $[\cdot]_+ = \max(0,\cdot)$; only the patient and visit
embeddings are updated. A patient is thereby placed according to the visits it
generates and a visit according to the concepts it records and its temporal
neighbors.

\subsection{Relation-Aware Message Passing}
\label{sec:mp}

The encoder $f_\theta$ maps the graph to a $d$-dimensional representation of each
visit through an input projection with temporal encoding, $L$ relation-aware
message-passing layers, and a cross-visit temporal attention, after which task
heads are trained jointly. Throughout, $i$ indexes the \emph{target} node and
$j$ a \emph{neighbor}; $H$ is the number of heads, $d_k=d/H$ the per-head width,
$[\,\cdot\,]_h$ the $h$-th $d_k$-block of a $d$-vector, $\|$ denotes concatenation, and $\sigma$ the logistic sigmoid.

\paragraph{Input projection and temporal encoding.}
Every node is first mapped into a common latent space, after which temporal
information is injected into visit nodes. Each node type receives its own affine
projection followed by a GELU nonlinearity,
\begin{equation}
\mathbf{z}^{(0)}_i = \mathrm{GELU}\!\bigl( \mathbf{W}^{\mathrm{in}}_{\phi(i)}\mathbf{x}_i + \mathbf{b}^{\mathrm{in}}_{\phi(i)} \bigr),
\label{eq:input}
\end{equation}
where $\mathbf{x}_i\in\mathbb{R}^{d_{\mathrm{in}}}$ is the pretrained feature,
$\mathbf{z}^{(0)}_i\in\mathbb{R}^{d}$ the projected latent, and
$\mathbf{W}^{\mathrm{in}}_{\tau}\in\mathbb{R}^{d\times d_{\mathrm{in}}}$,
$\mathbf{b}^{\mathrm{in}}_{\tau}$ are the type-dependent projection and bias parameters. Visit nodes are then displaced by a sinusoidal encoding of their admission time. For $j=0,\ldots,d/2-1$,
\begin{equation}
[\mathrm{PE}(t)]_j =
\begin{cases}
\sin\!\left(\dfrac{10^4 t}{10000^{2j/d}}\right), & \text{if $j$ is even}, \\[15pt]
\cos\!\left(\dfrac{10^4 t}{10000^{2j/d}}\right), & \text{if $j$ is odd},
\end{cases}
\label{eq:pe}
\end{equation}
we set $\mathbf{z}^{(0)}_i \leftarrow \mathbf{z}^{(0)}_i + \mathrm{PE}(t_i)$ for
every visit, where $\mathrm{PE}(t)\in\mathbb{R}^d$ is the deterministic code for
time $t$; the factor $10^4$ rescales $t\in[0,1]$ across the full observation window.

Each layer $\ell$ maps $\{\mathbf{z}^{(\ell)}_i\}$ to $\{\mathbf{z}^{(\ell+1)}_i\}$
in two stages: messages are aggregated \emph{within} each relation, then the
per-relation summaries are combined \emph{across} relations. 

\paragraph{Stage 1: intra-relation attention.}
This computes, for each relation separately, how much a node attends to its
neighbors under that relation. Queries are projected by the target type and keys
and values by the neighbor type,
\begin{equation}
\begin{aligned}
\mathbf{q}^h_i &= \bigl[\mathbf{W}^{Q}_{\phi(i)}\mathbf{z}^{(\ell)}_i\bigr]_h,
\qquad
\mathbf{k}^h_j = \bigl[\mathbf{W}^{K}_{\phi(j)}\mathbf{z}^{(\ell)}_j\bigr]_h, \\
&\hspace{2cm}
\mathbf{v}^h_j = \bigl[\mathbf{W}^{V}_{\phi(j)}\mathbf{z}^{(\ell)}_j\bigr]_h,
\end{aligned}
\label{eq:qkv}
\end{equation}
where $\mathbf{q}^h_i,\mathbf{k}^h_j,\mathbf{v}^h_j\in\mathbb{R}^{d_k}$ are the
query, key, and value of head $h$. Each relation $r$ owns per-head matrices
$\mathbf{M}^{h}_{r},\mathbf{R}^{h}_{r}\in\mathbb{R}^{d_k\times d_k}$ and a scalar
$\mu^{h}_{r}$. For an edge $j\!\to\!i$ 
of type $r$,
\begin{equation}
\alpha^{r,h}_{ij} = \frac{\exp(e^{r,h}_{ij})}{\sum\limits_{j'\in\mathcal{N}_r(i)} \exp(e^{r,h}_{ij'})},
\quad
e^{r,h}_{ij} = \frac{\mu^{h}_{r}}{\sqrt{d_k}}\,\bigl\langle \mathbf{q}^h_i,\, \mathbf{M}^{h}_{r}\mathbf{k}^h_j \bigr\rangle,
\label{eq:attn1}
\end{equation}
normalized over the relation-$r$ neighborhood $\mathcal{N}_r(i)$ \emph{alone}.
Here $\mathbf{M}^{h}_{r}$ transforms the key so that query--key matching is
relation-specific (the same neighbor is attended to differently depending on the
relation connecting it), and $\mu^{h}_{r}$ is a learned scalar that scales
relation $r$'s attention before the softmax, letting the model emphasize
clinically informative relations and suppress others. The per-relation summary
aggregates the values and applies a target-type output projection
$\mathbf{A}_{\phi(i)}\in\mathbb{R}^{d\times d}$,
\begin{equation}
\mathbf{z}^{(\ell)}_{i,r} = \mathbf{A}_{\phi(i)}
\Bigl(\,\big\|_{h=1}^{H} \textstyle\sum\limits_{j\in\mathcal{N}_r(i)} \alpha^{r,h}_{ij}\,\mathbf{R}^{h}_{r}\mathbf{v}^h_j \Bigr),
\label{eq:stage1}
\end{equation}
where $\mathbf{R}^{h}_{r}$ transforms the message so that the content passed
depends on the relation type, and $\mathbf{z}^{(\ell)}_{i,r}\in\mathbb{R}^{d}$ is
node $i$'s representation under relation $r$.

\paragraph{Stage 2: inter-relation attention.}
This decides which relations matter for a node, given its per-relation summaries. Let $\mathcal{R}(i)=\{r:\mathcal{N}_r(i)\neq\emptyset\}$ be the relations delivering to $i$. Just as stage 1 scores neighbors against a node's query, stage 2 scores each relation's summary against a learned \emph{type-level} query
$\mathbf{u}_{\phi(i)}\in\mathbb{R}^{d}$, and combines them accordingly,
\begin{equation}
\begin{aligned}
 \hat{\mathbf{z}}^{(\ell)}_i &= \sum_{r\in\mathcal{R}(i)} \beta^{r}_i\, \mathbf{z}^{(\ell)}_{i,r}, \\
\beta^{r}_i &= \frac{\exp\bigl(\langle \tanh(\mathbf{W}^{S}_{r}\mathbf{z}^{(\ell)}_{i,r}),\, \mathbf{u}_{\phi(i)}\rangle\bigr)}
{\sum\limits_{r'\in\mathcal{R}(i)} \exp\bigl(\langle \tanh(\mathbf{W}^{S}_{r'}\mathbf{z}^{(\ell)}_{i,r'}),\, \mathbf{u}_{\phi(i)}\rangle\bigr)},   
\end{aligned}
\label{eq:stage2}
\end{equation}
where $\beta^{r}_i\in\mathbb{R}$ is the weight of relation $r$ and
$\mathbf{W}^{S}_{r}\in\mathbb{R}^{d\times d}$ a per-relation scoring transform.
The transform and nonlinearity enter only the \emph{scoring}; the quantity mixed
is the untransformed summary $\mathbf{z}^{(\ell)}_{i,r}$. As $\beta^{r}_i$ is
per node, two visits may weight the same relation differently.

\paragraph{Gated residual.}
The aggregate $\hat{\mathbf{z}}^{(\ell)}_i$ is fused with the layer input through
a per-type gate. Each type carries a scalar $\omega_\tau$ and residual map
$\mathbf{W}^{R}_{\tau}$; with $\eta_\tau=\sigma(\omega_\tau)$,
\begin{equation}
\mathbf{z}^{(\ell+1)}_i \!\!= \mathrm{Dropout}\Bigl(\mathrm{LN}\bigl(
\eta_{\phi(i)}\,\hat{\mathbf{z}}^{(\ell)}_i + (1-\eta_{\phi(i)})\,\mathbf{W}^{R}_{\phi(i)}\mathbf{z}^{(\ell)}_i \bigr)\Bigr),
\label{eq:residual}
\end{equation}
where $\mathrm{LN}$ denotes layer norm and $\mathbf{z}^{(\ell+1)}_i=\mathbf{z}^{(\ell)}_i$ when $\mathcal{R}(i)=\emptyset$.
The gate $\eta_{\phi(i)}$ controls the update strength, so a type whose
neighborhood is uninformative can preserve its representation. We write
$\mathbf{z}_i\equiv\mathbf{z}^{(L)}_i$ for the final-layer output.

\paragraph{Cross-visit temporal attention.}
Message passing propagates history only one visit per layer while temporal attention gives every visit direct access to its entire history in a single step. Applied once after the final layer, it takes the message-passing output
$\mathbf{z}_i \equiv \mathbf{z}^{(L)}_i$ and, over its predecessors
$\mathcal{P}(i)$ (Def.~\ref{def:httg}), produces a refined representation
$\tilde{\mathbf{z}}_i$,
\begin{equation}
\begin{aligned}
    \tilde{\mathbf{z}}_i &= \mathrm{LN}\!\Bigl( \mathbf{z}_i + \mathbf{W}_{O}\,\mathrm{ReLU}\!\bigl( \textstyle\sum\limits_{j\in\mathcal{P}(i)} \gamma_{ij}\,\mathbf{W}_{V}\mathbf{z}_j \bigr) \Bigr),\\
\gamma_{ij} &= \frac{\exp\bigl(\langle \mathbf{W}_{Q}\mathbf{z}_i,\, \mathbf{W}_{K}\mathbf{z}_j\rangle / \sqrt{d}\bigr)}
{\sum\limits_{j'\in\mathcal{P}(i)} \exp\bigl(\langle \mathbf{W}_{Q}\mathbf{z}_i,\, \mathbf{W}_{K}\mathbf{z}_{j'}\rangle / \sqrt{d}\bigr)},
\end{aligned}
\label{eq:tempattn}
\end{equation}
where $\mathbf{z}_i$ is the message-passing output of the current visit,
$\gamma_{ij}\in\mathbb{R}$ its attention weight on predecessor $j$,
$\mathbf{W}_{Q},\mathbf{W}_{K},\mathbf{W}_{V},\mathbf{W}_{O}\in\mathbb{R}^{d\times d}$
learnable projections, and $\tilde{\mathbf{z}}_i$ the temporally refined
representation; $\tilde{\mathbf{z}}_i=\mathbf{z}_i$ when $\mathcal{P}(i)=\emptyset$,
as for a first admission. Since $\mathcal{P}(i)$ holds only strictly earlier
visits, the step is causal by construction.

\paragraph{Task heads.}
The shared representation $\tilde{\mathbf{z}}_i$ is read by one affine head per
task $k\in\mathcal{T}$, $\hat{\mathbf{y}}^{k}_i = \mathbf{W}^{k}\tilde{\mathbf{z}}_i + \mathbf{b}^{k}$,
$\mathbf{W}^{k}\in\mathbb{R}^{|\mathcal{Y}_k|\times d}$. Mortality and readmission
use binary cross-entropy over two classes, length of stay cross-entropy over its
ten buckets, and drug recommendation binary cross-entropy over the prescription
vocabulary $\mathcal{M}$ on a temperature-scaled logit. We write $\mathcal{L}_k$
for the resulting per-task loss of Problem~\ref{prob:mtl}.

\paragraph{Dual-balancing objective.}
Rather than balancing the losses $\mathcal{L}_{k}$ for each task directly, DB-MTL~\cite{lin2023dbmtl} balances their \emph{gradients}, preventing any single task from dominating the shared encoder. This realizes the combiner $\Phi$ of Problem~\ref{prob:mtl} at the gradient level. Let $\theta$ be the encoder
parameters shared by all tasks and $\varphi_k$ the parameters of head $k$. Loss
\emph{scale} is removed by differentiating the logarithm of each loss, and the
resulting gradient is smoothed across steps by an exponential moving average
with momentum $\rho\in[0,1)$,
\begin{equation}
\bar{\mathbf{g}}_k^{(t)} = \rho\,\bar{\mathbf{g}}_k^{(t-1)}\!\! + (1-\rho)\,\mathbf{g}_k^{(t)},
\quad
\mathbf{g}_k^{(t)} = \nabla_{\theta}\log(\mathcal{L}_k^{(t)} + \varepsilon),
\label{eq:dbmtlema}
\end{equation}
where $\mathbf{g}_k$ is task $k$'s current log-loss gradient, $\bar{\mathbf{g}}_k$ its running average (a convex combination of the previous average and the current gradient, i.e.\ an exponential moving average), $t$ denotes training step, and $\varepsilon$ a small constant. Gradient \emph{magnitude} is then equalized by rescaling every task to
the largest running norm and summing the unit directions,
\begin{equation}
\mathbf{g}^{\mathrm{shared}} = \max_{k\in\mathcal{T}} \lVert \bar{\mathbf{g}}_k \rVert_2 \!\sum_{k\in\mathcal{T}}\! \frac{\bar{\mathbf{g}}_k}{\lVert \bar{\mathbf{g}}_k \rVert_2 + \varepsilon},
\label{eq:dbmtlcombine}
\end{equation}
The shared parameters $\theta$ are updated with $\mathbf{g}^{\mathrm{shared}}$, so every task contributes a direction of equal length, while each head is updated
with its own gradient $\nabla_{\varphi_k}\log(\mathcal{L}_k+\varepsilon)$,
unaffected by the other tasks.

\section{Experiments}
\label{sec:experiments}
\begin{table*}[t]
\centering
\footnotesize
\caption{Mortality, readmission, and length-of-stay (10-class) prediction on MIMIC-III and MIMIC-IV. Mortality and readmission report AUROC and AUPR; length-of-stay reports Accuracy, AUROC, and $F_1$ (all in \%). Best per column in bold, second best underlined.}
\label{tab:results_combined}
\setlength{\tabcolsep}{2pt}
\begin{tabular}{l@{\hspace{3pt}}cccc cccc cccccc}
\toprule
\multirow{3}{*}{Model}
  & \multicolumn{4}{c}{Mortality}
  & \multicolumn{4}{c}{Readmission}
  & \multicolumn{6}{c}{Length-of-stay} \\
\cmidrule(lr){2-5} \cmidrule(lr){6-9} \cmidrule(lr){10-15}
  & \multicolumn{2}{c}{MIMIC-III} & \multicolumn{2}{c}{MIMIC-IV}
  & \multicolumn{2}{c}{MIMIC-III} & \multicolumn{2}{c}{MIMIC-IV}
  & \multicolumn{3}{c}{MIMIC-III} & \multicolumn{3}{c}{MIMIC-IV} \\
\cmidrule(lr){2-3} \cmidrule(lr){4-5} \cmidrule(lr){6-7} \cmidrule(lr){8-9} \cmidrule(lr){10-12} \cmidrule(lr){13-15}
  & AUROC & AUPR & AUROC & AUPR
  & AUROC & AUPR & AUROC & AUPR
  & Acc. & AUROC & $F_1$ & Acc. & AUROC & $F_1$ \\
\midrule
GRU
  & 61.09 & 9.23 & 62.95 & 3.00 & 65.00 & 66.47 & 68.30 & 69.10
  & 41.32 & 80.53 & $\mathbf{36.46}$ & 36.91 & 79.93 & 33.11 \\
Transformer
  & 60.60 & 9.62 & 68.53 & 3.68 & 64.34 & 65.84 & 68.90 & 69.74
  & 41.23 & 79.99 & 35.76 & 37.35 & 80.73 & 32.38 \\
Deepr
  & 58.61 & \underbar{$11.87$} & 65.13 & 3.20 & 65.10 & 68.68 & 67.20 & 68.06
  & 39.31 & 78.02 & 25.09 & 36.00 & 80.53 & 31.05 \\
GRAM
  & 60.00 & 11.00 & 65.00 & 4.00 & 64.00 & 67.00 & 66.00 & 66.00
  & 40.00 & 78.00 & 29.00 & 35.00 & 79.00 & 32.00 \\
Concare
  & 59.21 & 9.43 & 49.90 & 1.76 & 60.46 & 61.41 & 50.00 & 51.09
  & 39.16 & 78.78 & 32.35 & 32.39 & 76.47 & 26.09 \\
Dr. Agent
  & 60.35 & 9.92 & 68.54 & 4.21 & 64.76 & 63.80 & 68.77 & 69.71
  & 41.75 & 80.30 & 35.64 & 37.96 & 80.98 & 33.77 \\
AdaCare
  & 59.13 & 10.35 & 67.03 & 4.02 & 63.82 & 65.62 & 68.96 & 69.87
  & 41.57 & \underbar{$80.73$} & \underbar{$36.32$} & 38.25 & $\mathbf{81.42}$ & 34.38 \\
StageNet
  & 59.69 & 11.75 & 63.19 & 3.04 & 63.83 & 65.61 & 68.94 & 69.80
  & 41.09 & 80.34 & 35.87 & 38.19 & \underbar{$81.07$} & 34.19 \\
GRASP
  & 59.29 & 9.32 & 49.99 & 1.77 & \underbar{$66.91$} & \underbar{$70.41$} & 50.55 & 51.37
  & 40.66 & 78.97 & 22.80 & 35.28 & 79.86 & 26.95 \\
MulT-EHR
  & \underbar{$62.38$} & 10.46 & \underbar{$78.79$} & \underbar{$6.85$} & 64.61 & 67.80 & \underbar{$76.89$} & \underbar{$77.40$}
  & $\mathbf{43.93}$ & $\mathbf{81.00}$ & 34.53 & $\mathbf{42.20}$ & 79.50 & $\mathbf{39.11}$ \\
\midrule
\textbf{MiGHT-EHR}
  & \textbf{75.54} & \textbf{14.68} & $\textbf{78.84}$ & $\textbf{8.71}$ & \textbf{80.19} & \textbf{86.88} & $\textbf{83.73}$ & $\textbf{85.496}$
  & \underbar{$42.22$} & 79.99 & 35.57 & \underbar{$39.16$} & 79.26 & \underbar{$38.88$} \\
\bottomrule
\end{tabular}
\end{table*}
\subsection{Settings}

\paragraph{Datasets.}
We evaluate on MIMIC-III \cite{johnson2016mimic} and MIMIC-IV \cite{johnson2023mimic}, the standard large-scale public EHR benchmarks, which allow fair comparison with prior work and test generalization across two independent hospital databases. Graph statistics, including node and edge counts are summarized in Appendix.

\paragraph{Tasks and Evaluation Metrics.}
We jointly train on four prediction tasks: mortality prediction (Mort) and readmission prediction (Readm) as binary classification, length of stay (LoS) as 10-class classification over duration buckets, and drug recommendation (Drug\_rec) as multi-label classification over $351$ labels on MIMIC-III and $556$ on MIMIC-IV. Performance is evaluated using AUROC, AUPR, accuracy, $F_1$, and Jaccard index, as appropriate. All methods use the same fixed $90/10$ visit-level split, and results are reported from the checkpoint maximizing the mean test AUROC across the four tasks. Metric definitions are provided in Appendix.

\subsection{Implementation Details}

MiGHT-EHR is implemented in \textit{PyTorch}, with \textit{DGL}~\cite{wang2019deep} for graph operations and \textit{PyHealth}~\cite{zhao2021pyhealth} for EHR processing and baseline benchmarking. Models are trained for $1000$ epochs using Adam with learning rates of $3\times10^{-4}$ (MIMIC-III) and $5\times10^{-4}$ (MIMIC-IV), weight decay $1\times10^{-5}$, and dropout $0.3$. Cross-entropy is used for the classification tasks and binary cross-entropy for drug recommendation, with task losses combined using DB-MTL ($\rho=0.9$). All experiments run on NVIDIA Quadro RTX 6000 GPUs ($24$ GB each), with a single GPU for training and additional GPUs for parallel evaluation.

\subsection{Comparable Methods}

We compare against sequential EHR models, including GRU~\cite{medsker2001RNN}, Transformer~\cite{vaswani2017transformer}, Deepr~\cite{nguyen2016deepr}, ConCare~\cite{ma2020concare}, Dr.~Agent~\cite{gao2020dragent}, AdaCare~\cite{ma2020adacare}, StageNet~\cite{gao2020stagenet}, and GRASP~\cite{zhang2021grasp}, together with graph-based and ontology-aware methods including GRAM~\cite{choi2017gram}, GraphCare~\cite{jiang2023graphcare}, and MulT-EHR~\cite{chan_multitask_ehr}. For drug recommendation, we additionally compare against task-specific methods MICRON~\cite{yang2021micron}, MoleRec~\cite{yang2023molerec}, and SafeDrug~\cite{yang2021safedrug}. Baseline descriptions are provided in the Appendix.

\subsection{Quantitative Results}
\begin{table}[t]
\centering
\footnotesize
\caption{Drug recommendation. Best per column in bold, second best underlined}
\label{tab:results_drug}
\setlength{\tabcolsep}{2pt}
\begin{tabular}{l@{\hspace{3pt}}cccccc}
\toprule
\multirow{2}{*}{Model}
  & \multicolumn{3}{c}{MIMIC-III} & \multicolumn{3}{c}{MIMIC-IV} \\
\cmidrule(lr){2-4} \cmidrule(lr){5-7}
  & AUROC & Jacc. & AUPR & AUROC & Jacc. & AUPR \\
\midrule
GRU                 & 94.85 & 34.12 & 67.69 & 95.69 & 25.58 & 61.27 \\
Transformer & 95.26 & 34.68 & 69.11 & 96.17 & 26.65 & 63.19 \\
GRAM      & 94.00 & $\mathbf{48.00}$ & $\mathbf{77.00}$ & 94.00 & $\mathbf{45.00}$ & 60.00 \\
Concare              & 94.74 & 31.03 & 67.12 & 95.69 & 25.58 & 61.23 \\
Deepr              & \underbar{$96.09$} & 44.45 & 62.48 & \underbar{$97.35$} & 43.31 & 60.31 \\
Dr. Agent           & 94.92 & 34.85 & 68.16 & 95.69 & 25.58 & 61.24 \\
AdaCare         & 94.75 & 30.97 & 67.07 & 95.69 & 25.58 & 61.26 \\
StageNet          & 94.82 & 33.52 & 67.91 & 95.71 & 25.76 & 61.35 \\
GRASP               & 96.01 & 44.12 & 62.53 & 95.69 & 25.57 & 61.26 \\
MulT-EHR        & 95.76 & 45.20 & \underbar{$72.04$} & $\mathbf{97.66}$ & 41.46 & $\mathbf{72.61}$ \\
\midrule
MICRON              & $\mathbf{96.21}$ & 45.95 & 63.84 & 95.68 & 25.58 & 61.19 \\
MoleRec         & 92.00 & 43.10 & 69.80 & 92.10 & 40.60 & 68.60 \\
SafeDrug & 94.20 & \underbar{$47.20$} & 69.40 & 91.80 & \underbar{$44.30$} & 66.40 \\
\midrule
\textbf{MiGHT-EHR} & 94.90 & 41.90 & 68.14 & 97.00 & 40.81 & \underbar{$68.75$} \\
\bottomrule
\end{tabular}
\end{table}
Tables~\ref{tab:results_combined}--\ref{tab:results_drug} report per-task results on both datasets, while Table~\ref{tab:delta_p} summarizes the average relative improvement over MulT-EHR. For method $\mathcal{A}$ and task $t$ with $N_t$ metrics pooled across both datasets,
\begin{equation}
\Delta_{p,t} \!= \frac{100}{N_t}\sum_{i=1}^{N_t}
\frac{M^{\mathcal{A}}_{t,i}-M^{\text{MulT-EHR}}_{t,i}}{M^{\text{MulT-EHR}}_{t,i}},
\quad
\overline{\Delta_p}\! = \frac{1}{|T_{\mathcal{A}}|}\!\sum_{t}\!\Delta_{p,t}.
\label{eq:deltap}
\end{equation}

MiGHT-EHR achieves the best performance on average across all four tasks ($\overline{\Delta_p}=+8.61\%$), the only positive average improvement among the fourteen methods despite using a \emph{single} shared model across all tasks. The gains are concentrated on the two rare-event tasks, mortality ($+22.16\%$) and readmission ($+17.90\%$). Performance remains competitive for drug recommendation staying within one AUROC point of specialized methods~\cite{yang2021micron, yang2021safedrug, yang2023molerec}, while length of stay matches state-of-the-art accuracy performance. These improvements arise from learning a unified representation across related clinical tasks, enabled by statistically grounded NPMI graph construction, temporal attention over longitudinal patient histories, and balanced multi-task optimization via DB-MTL.

\begin{table}[t]
\centering
\footnotesize
\caption{Average relative improvement $\overline{\Delta_p}$ (\%) over MulT-EHR,
pooled across MIMIC-III and MIMIC-IV.}
\label{tab:delta_p}
\setlength{\tabcolsep}{2pt}
\begin{tabular}{l@{\hspace{3pt}}ccccc}
\toprule
Method & Mort. & Readm. & Drug\_rec & LoS & $\overline{\Delta_p}$ \\
\midrule
GRU         & $-22.53$ &  $-5.81$ & $-14.57$ &  $-4.71$ & $-11.91$ \\
Transformer & $-17.55$ &  $-5.90$ & $-13.01$ &  $-5.16$ & $-10.41$ \\
Deepr       & $-15.80$ &  $-5.65$ &  $-4.56$ & $-12.59$ &  $-9.65$ \\
GRAM        & $-14.44$ &  $-7.75$ &  $-0.22$ & $-10.76$ &  $-8.29$ \\
Concare     & $-31.48$ & $-21.20$ & $-15.87$ & $-13.38$ & $-20.48$ \\
Dr. Agent   & $-14.99$ &  $-6.54$ & $-14.19$ &  $-4.08$ &  $-9.95$ \\
AdaCare     & $-15.63$ &  $-6.12$ & $-15.90$ &  $-3.26$ & $-10.23$ \\
StageNet    & $-16.85$ &  $-6.15$ & $-14.65$ &  $-3.92$ & $-10.39$ \\
GRASP       & $-31.64$ & $-15.12$ & $-11.88$ & $-15.16$ & $-18.45$ \\
MICRON      &    --    &    --    & $-10.89$ &    --    & $-10.89$ \\
MoleRec     &    --    &    --    &  $-4.16$ &    --    &  $-4.16$ \\
SafeDrug    &    --    &    --    &  $-1.43$ &    --    &  $-1.43$ \\
MulT-EHR    &  $0.00$  &  $0.00$  &  $0.00$  &  $0.00$  &  $0.00$  \\
\midrule
\textbf{MiGHT-EHR} & $\mathbf{+22.16}$ & $\mathbf{+17.90}$ & $-3.53$ & $-1.70$ & $\mathbf{+8.71}$ \\
\bottomrule
\end{tabular}
\end{table}

\subsection{Qualitative Evaluation}

To explain these quantitative gains, we analyze the frozen MIMIC-III (MIMIC-IV results are in Appendix) model without further training. All results are computed on the held-out test split by examining visit representations at four stages: $\mathrm{L}_{in}$ (pretrained input), $L_0$ (after input projection and temporal encoding), and $L_1,L_2$ (after the two message-passing layers), allowing us to isolate the contribution of graph representation learning.

\begin{figure}[tb]
\centering
\begin{tabular}{@{}c@{\hspace{0.02\columnwidth}}c@{}}
\raisebox{0.013\columnwidth}{%
  \includegraphics[width=0.45\columnwidth,height=0.38\columnwidth]%
  {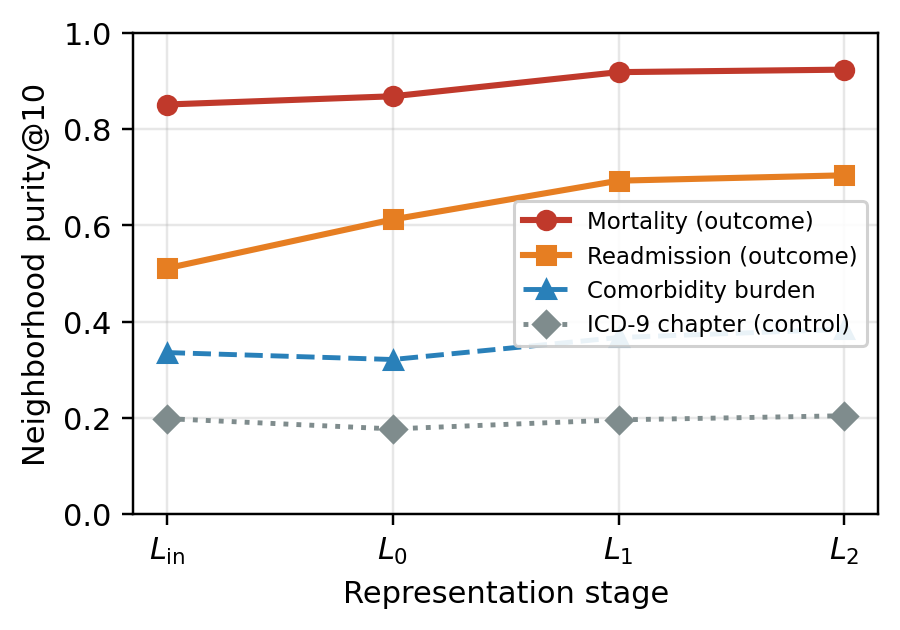}}
&
\includegraphics[width=0.52\columnwidth]{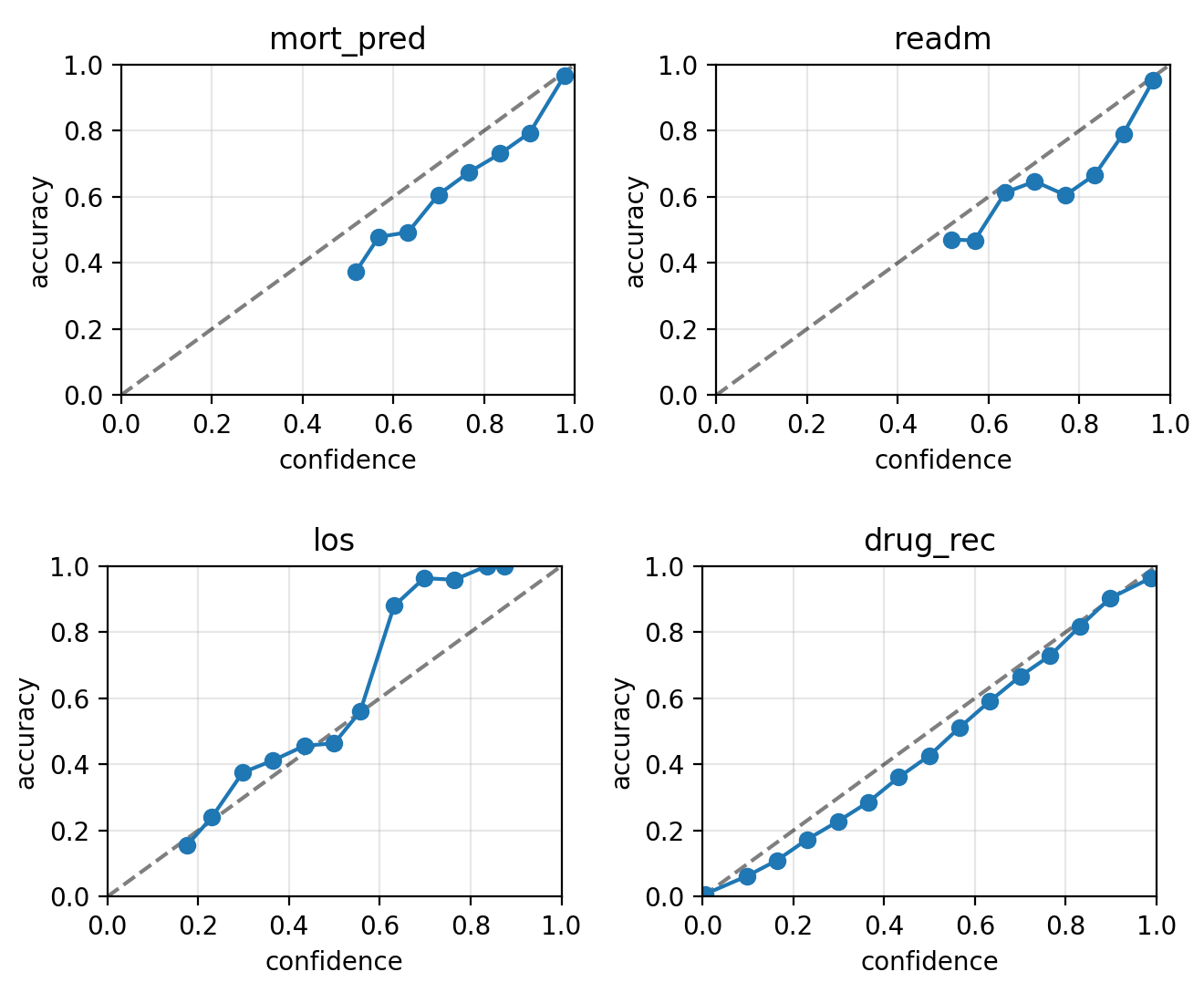}
\end{tabular}
\caption{Post-hoc analysis on frozen MIMIC-III. Left: neighborhood purity across
stages. Right: test reliability; drug and LoS are well calibrated, while
rare-event outcomes are mildly overconfident.}
\label{fig:posthoc_analysis}
\end{figure}

\paragraph{Neighborhoods are organized by outcome.}
Figure~\ref{fig:posthoc_analysis} measures how often a visit's $10$ nearest neighbors share its label at each network stage; since the labels differ in class count, only the trend with depth is meaningful. Neighbors increasingly share the same clinical outcome with depth ($0.924$ for mortality, $0.704$ for readmission at $L_2$), whereas comorbidity burden rises only mildly and diagnostic chapter stays flat. These two serve as controls: the representation organizes visits by \emph{prognosis} rather than diagnosis, ruling out that outcome clustering merely reflects diagnostic similarity

\paragraph{Clinical concepts are linear directions.}
To assess whether clinical concepts are encoded in the representation, we fit a linear probe to each concept group at $L_2$ (Table~\ref{tab:cav}). Several concepts are highly recoverable, led by circulatory-system (AUROC $0.785$), whereas others remain near chance. Recoverability follows clinical salience rather than frequency: injury and poisoning, the most frequent group ($15{,}650$ visits), achieves only $0.604$, while the much rarer perinatal group reaches $0.715$. The representation therefore captures clinically meaningful concepts rather than simply reflecting concept frequency.

\begin{table}[tb]
\centering
\small
\caption{Concept-probe AUROC at $L_2$ (frozen MIMIC-III).}
\label{tab:cav}
\begin{tabular}{@{}lrc@{}}
\toprule
Concept group & $n$ & Probe AUROC \\
\midrule
Circulatory system          & 3{,}482  & \textbf{0.785} \\
Perinatal conditions        & 826      & 0.715 \\
Endocrine / metabolic       & 472      & 0.672 \\
Infectious \& parasitic     & 955      & 0.671 \\
Digestive system            & 4{,}753  & 0.668 \\
Neoplasms                   & 4{,}090  & 0.633 \\
Injury \& poisoning         & 15{,}650 & 0.604 \\
Respiratory system          & 270      & 0.535 \\
Pregnancy \& childbirth     & 372      & 0.526 \\
\bottomrule
\end{tabular}
\end{table}

\paragraph{Predicted probabilities are calibrated.}
Figure~\ref{fig:posthoc_analysis} show that drug recommendation is well calibrated (ECE (expected calibration error) = $0.013$), with length of stay remaining reasonably calibrated (ECE = $0.037$). Mortality and readmission are mildly overconfident (ECE  = $0.085$ and $0.103$ respectively), a common characteristic of low-prevalence outcomes that can be further improved through post-hoc calibration. Overall, MiGHT-EHR produces predictions that are both well ranked and well calibrated, an important property for clinical decision support.


\section{Conclusion and Future Work}
\label{sec:conclusion}

We presented MiGHT-EHR, a heterogeneous graph transformer with temporal attention for multi-task clinical prediction. It achieves the best performance on average across four tasks on MIMIC-III and MIMIC-IV datasets. Two extensions are especially promising: first, modeling the concepts within a visit as true hyperedges would capture their joint, higher-order interaction, which our pairwise association edges only approximate; second, letting the graph structure evolve over time would model temporal dynamics beyond the visit ordering our static graph already captures.

\appendix

\begin{center}
{\Large\bfseries Technical Appendix\par}
\end{center}

\noindent
This appendix provides supplementary material for the main paper. It details the
concept vocabulary and CCS mapping (Appendix~\ref{app:vocab}), the full dataset
and graph statistics (Appendix~\ref{app:graph_stats}), the evaluation metrics
(Appendix~\ref{app:metrics}), the baseline methods (Appendix~\ref{app:baselines}),
the hyperparameters and training details (Appendix~\ref{app:hyperparams}), and a
post-hoc analysis of the frozen MIMIC-IV model (Appendix~\ref{app:posthoc_mimic4}).


\section{Concept Vocabulary and CCS Mapping}
\label{app:vocab}

Diagnosis and procedure nodes are represented at the level of Clinical
Classifications Software (CCS) categories~\cite{hcup_ccs} rather than raw
International Classification of Diseases (ICD) codes. Prescription nodes use
normalized drug codes and are left unmapped.

\paragraph{Why CCS.}
Raw ICD codes fragment a single clinical concept in two ways. First, ICD is
highly granular: many distinct codes denote clinically equivalent conditions,
so a concept is split across several sparsely-observed nodes. Second, and more
consequentially for MIMIC-IV, the database spans the ICD-9 to ICD-10 transition,
during which the same condition is recorded under different code systems before
and after the switch; a raw-code graph therefore represents one condition as two
disconnected nodes. Mapping to CCS resolves both: each ICD code is assigned to
one of a few hundred CCS categories, unifying ICD-9 and ICD-10 variants of a
condition into a single node. This matters not only for representation but for
the association estimates based on normalized pointwise
mutual information: this measure requires each concept to be observed often
enough for its marginal
$p(a)$ and joint $p(a,b)$ to be estimated reliably, and CCS categories are
observed frequently enough where individual ICD codes are not.

\paragraph{Mapping.}
Diagnoses are mapped to CCS Clinical Modification (CCSCM) and procedures to CCS
Procedure categories (CCSPROC), with both ICD-9 and ICD-10 sources mapped into
the same category space:
\begin{center}
\small
\begin{tabular}{ll}
\toprule
Source code system & Target category \\
\midrule
ICD-9-CM, ICD-10-CM     & CCSCM (diagnosis) \\
ICD-9-PCS, ICD-10-PCS   & CCSPROC (procedure) \\
NDC (drugs)             & unmapped (normalized drug code) \\
\bottomrule
\end{tabular}
\end{center}

\paragraph{Resulting vocabularies.}
The mapping reduces the diagnosis and procedure vocabularies by one to two
orders of magnitude while leaving the cohort unchanged
(Table~\ref{tab:dataset_stats}). On MIMIC-III, $6{,}984$ ICD-9 diagnosis codes
map to $281$ CCS categories and $2{,}032$ procedure codes to $221$, with no
unmapped codes. On MIMIC-IV, where ICD-9 and ICD-10 coexist, $28{,}562$
diagnosis codes map to $280$ categories and $14{,}911$ procedure codes to $231$;
the far larger raw vocabulary reflects the dual coding systems, and the near-
identical category counts across the two databases ($281$ vs $280$ diagnoses,
$221$ vs $231$ procedures) confirm that CCS places both on a common concept
space. Prescriptions remain at the normalized-drug-code level, with $4{,}204$
codes on MIMIC-III and $6{,}588$ on MIMIC-IV.

\section{Dataset and Graph Statistics}
\label{app:graph_stats}

Table~\ref{tab:dataset_stats} summarizes the node types, their counts, and the
number of labeled observations per task on both databases.
Table~\ref{tab:graph_stats} reports the full graph structure, including the
membership, temporal, and co-occurrence edge counts.
Table~\ref{tab:npmi} reports the effect of the association criterion, showing
the fraction of observed concept pairs retained under
$\mathrm{NPMI}\geq\tau$ and $c(a,b)\geq\kappa$.
Table~\ref{tab:temporal_stats} reports the temporal structure of the patient
trajectories, over which the cross-visit temporal attention operates; MIMIC-IV is
markedly more longitudinal than MIMIC-III.

\begin{table}[t]
\centering
\small
\setlength{\tabcolsep}{5pt}
\caption{Dataset and graph statistics. For diagnoses and procedures,
\emph{Count} shows the number of raw ICD codes $\to$ the number of CCS
categories used as graph nodes; prescriptions are NDC-level and unmapped.
\emph{Avg./Visit} is the mean number of concepts of that type recorded in a
visit. \emph{No.\ Obs.} is the number of labelled visits for each task. The
graphs also contain directed temporal edges linking each patient's consecutive
visits ($12{,}456$ for MIMIC-III, $322{,}576$ for MIMIC-IV); co-occurrence edge
statistics are reported in Table~\ref{tab:npmi}.}
\label{tab:dataset_stats}
\begin{tabular}{lrr@{\hskip 10pt}lr}
\toprule
\multicolumn{5}{c}{\textbf{MIMIC-III}} \\
\cmidrule(lr){1-3} \cmidrule(lr){4-5}
Node Type & Count & Avg./Visit & Task & No.\ Obs. \\
\midrule
Patients      & 46{,}520          & ---   & Mortality    & 9{,}718 \\
Visits        & 58{,}976          & ---   & Readmission  & 9{,}718 \\
Diagnoses     & 6{,}984\,$\to$\,281  & 9.49  & LoS          & 44{,}407 \\
Prescriptions & 4{,}204           & 33.53 & Drug Recomm. & 14{,}142 \\
Procedures    & 2{,}032\,$\to$\,221  & 3.07  &              & \\
\midrule
\multicolumn{5}{c}{\textbf{MIMIC-IV}} \\
\cmidrule(lr){1-3} \cmidrule(lr){4-5}
Node Type & Count & Avg./Visit & Task & No.\ Obs. \\
\midrule
Patients      & 223{,}452           & ---   & Mortality    & 160{,}388 \\
Visits        & 546{,}028           & ---   & Readmission  & 160{,}388 \\
Diagnoses     & 28{,}562\,$\to$\,280   & 9.91  & LoS          & 277{,}438 \\
Prescriptions & 6{,}588             & 21.71 & Drug Recomm. & 187{,}224 \\
Procedures    & 14{,}911\,$\to$\,231   & 1.25  &              & \\
\bottomrule
\end{tabular}
\end{table}

\begin{table}[t]
\centering
\small
\caption{Statistics of the constructed graphs. Both databases use the same
concept vocabulary (CCS categories for diagnoses and procedures, NDC codes for
drugs) and the same association criterion, $\mathrm{NPMI} \geq \tau = 0.10$ and
$c(a,b) \geq \kappa = 5$. Membership counts are unique visit--concept pairs.
Same-type co-occurrence relations are stored symmetrically, so their edge counts
are twice the number of retained concept pairs.}
\label{tab:graph_stats}
\begin{tabular}{lrr}
\toprule
 & \textbf{MIMIC-III} & \textbf{MIMIC-IV} \\
\midrule
\multicolumn{3}{l}{\textit{Nodes}} \\
patient    & 46{,}520  & 223{,}452 \\
visit      & 58{,}976  & 546{,}028 \\
diagnosis (CCSCM)   & 281 & 280 \\
procedure (CCSPROC) & 221 & 231 \\
drug (NDC) & 4{,}204   & 6{,}588 \\
\midrule
\multicolumn{3}{l}{\textit{Membership relations}} \\
\texttt{makes}      & 58{,}976    & 546{,}028 \\
\texttt{diagnosed}  & 559{,}963   & 5{,}410{,}351 \\
\texttt{treated}    & 181{,}334   & 681{,}516 \\
\texttt{prescribed} & 1{,}977{,}710 & 11{,}853{,}298 \\
\midrule
\multicolumn{3}{l}{\textit{Temporal relation}} \\
\texttt{next\_visit} & 12{,}456 & 322{,}576 \\
\midrule
\multicolumn{3}{l}{\textit{Co-occurrence relations (NPMI)}} \\
\texttt{co\_diag}       & 7{,}762   & 7{,}094 \\
\texttt{co\_proc}       & 3{,}982   & 6{,}358 \\
\texttt{co\_drug}       & 629{,}712 & 1{,}815{,}272 \\
\texttt{co\_diag\_proc} & 4{,}084   & 5{,}312 \\
\texttt{co\_diag\_drug} & 54{,}794  & 108{,}813 \\
\texttt{co\_proc\_drug} & 36{,}542  & 81{,}149 \\
\midrule
\multicolumn{3}{l}{\textit{Concepts per visit (mean)}} \\
diagnoses  & 9.5  & 9.9 \\
procedures & 3.1  & 1.2 \\
drugs      & 33.5 & 21.7 \\
\bottomrule
\end{tabular}
\end{table}

\begin{table}[t]
\centering
\small
\setlength{\tabcolsep}{5pt}
\caption{Effect of the association criterion on co-occurrence edges.
\emph{Observed} counts concept pairs appearing together in at least one visit;
\emph{Retained} counts those passing $\mathrm{NPMI} \geq \tau = 0.10$ and
$c(a,b) \geq \kappa = 5$. The criterion discards $75\%$--$90\%$ of observed
pairs at comparable rates on both databases, keeping only statistically
associated ones.}
\label{tab:npmi}
\begin{tabular}{lrrr}
\toprule
Concept pair & Observed & Retained & \% kept \\
\midrule
\multicolumn{4}{c}{\textit{MIMIC-III}} \\
\cmidrule(lr){1-4}
Diagnosis--Diagnosis     & 27{,}666      & 3{,}881   & 14.0 \\
Procedure--Procedure     & 9{,}445       & 1{,}991   & 21.1 \\
Prescription--Prescription & 1{,}537{,}572 & 314{,}856 & 20.5 \\
Diagnosis--Procedure     & 33{,}445      & 4{,}084   & 12.2 \\
Diagnosis--Prescription  & 482{,}648     & 54{,}794  & 11.4 \\
Procedure--Prescription  & 260{,}539     & 36{,}542  & 14.0 \\
\midrule
\multicolumn{4}{c}{\textit{MIMIC-IV}} \\
\cmidrule(lr){1-4}
Diagnosis--Diagnosis     & 34{,}725      & 3{,}547   & 10.2 \\
Procedure--Procedure     & 13{,}272      & 3{,}179   & 24.0 \\
Prescription--Prescription & 3{,}645{,}169 & 907{,}636 & 24.9 \\
Diagnosis--Procedure     & 48{,}181      & 5{,}312   & 11.0 \\
Diagnosis--Prescription  & 1{,}044{,}545 & 108{,}813 & 10.4 \\
Procedure--Prescription  & 528{,}633     & 81{,}149  & 15.4 \\
\bottomrule
\end{tabular}
\end{table}

\begin{table}[t]
\centering
\small
\setlength{\tabcolsep}{6pt}
\caption{Temporal structure of the patient trajectories. A patient's visits are
linked in admission-time order by directed \texttt{next\_visit} edges, over
which the model attends. MIMIC-IV is substantially more longitudinal: nearly
half of its patients have repeat admissions and over three quarters of its
visits belong to a multi-visit trajectory, against one sixth and one third in
MIMIC-III.}
\label{tab:temporal_stats}
\begin{tabular}{l@{\hspace{-1pt}}r@{\hspace{6pt}}r}
\toprule
 & \textbf{MIMIC-III} & \textbf{MIMIC-IV} \\
\midrule
Visits per patient (mean)      & 1.27      & 2.44 \\
Visits per patient (max)       & 42        & 238 \\
\midrule
Patients with $\geq 2$ visits  & 7{,}537 \, (16.2\%)  & 100{,}163 \, (44.8\%) \\
\quad with $\geq 3$ visits     & 2{,}377 \, (5.1\%)   & 56{,}040 \, (25.1\%) \\
\quad with $\geq 5$ visits     & 527 \, (1.1\%)       & 24{,}760 \, (11.1\%) \\
\midrule
(\texttt{next\_visit}) edges & 12{,}456 & 322{,}576 \\
Visits in multi-visit trajectories    & 33.9\%   & 77.4\% \\
\bottomrule
\end{tabular}
\end{table}

\section{Evaluation Metrics}
\label{app:metrics}

We report standard ranking, classification, and calibration metrics. Consider
$N$ test visits; for a binary task, $y_i\in\{0,1\}$ is the label of visit $i$ and
$\hat{p}_i$ its predicted positive probability, and $\hat{y}_i$ the predicted
class.

\paragraph{AUROC.}
The area under the receiver-operating-characteristic curve equals the
probability that a randomly chosen positive is ranked above a randomly chosen
negative,
\begin{equation}
\mathrm{AUROC} = \Pr\bigl(\hat{p}_i > \hat{p}_j \mid y_i = 1,\, y_j = 0\bigr).
\end{equation}
For the multi-class length-of-stay task we report the macro one-versus-one
AUROC, averaging the binary AUROC over all pairs of classes; for the multi-label
drug-recommendation task we report the sample-averaged AUROC, computing AUROC
over each visit's label set and averaging across visits.

\paragraph{AUPR.}
The area under the precision--recall curve,
$\mathrm{AUPR} = \sum_{n}(R_n - R_{n-1})\,P_n$, where $P_n$ and $R_n$ are the
precision and recall at the $n$-th decision threshold. AUPR is more informative
than AUROC under heavy class imbalance, as in mortality prediction.

\paragraph{Accuracy.}
The fraction of visits whose predicted class matches the true class,
$\mathrm{Acc} = \frac{1}{N}\sum_{i} \mathbf{1}[\hat{y}_i = y_i]$. We report it for
the ten-class length-of-stay task.

\paragraph{F1 score.}
The harmonic mean of precision $P$ and recall $R$,
$F_1 = 2PR/(P+R)$. For the multi-class and multi-label tasks we report the
\emph{weighted} $F_1$, averaging the per-class $F_1$ weighted by class support.

\paragraph{Jaccard.}
For multi-label drug recommendation, the Jaccard index between the predicted
medication set $\hat{Y}$ and the true set $Y$,
$J = |\hat{Y}\cap Y| \,/\, |\hat{Y}\cup Y|$, taking the top-$k$ predicted
medications as $\hat{Y}$ and averaging over visits.

\paragraph{Expected Calibration Error (ECE).}
The \emph{confidence} of a prediction is the probability the model assigns to the
class it predicts, $\hat{c}_i = \max_{c}\hat{p}_{ic}$, where $\hat{p}_{ic}$ is the
predicted probability that visit $i$ belongs to class $c$ (for a binary task this
is the predicted probability of the chosen class). A model is calibrated when its
confidence matches its accuracy, that is, among all predictions made with
confidence $q$ a fraction $q$ are correct. ECE measures the average gap between
the two. We partition $[0,1]$ into $M$ equal-width bins by predicted confidence;
with $\mathrm{acc}(B_m)$ the accuracy and $\mathrm{conf}(B_m)$ the mean confidence
of the predictions in bin $B_m$,
\begin{equation}
\mathrm{ECE} = \sum_{m=1}^{M} \frac{|B_m|}{N}\,\bigl|\,\mathrm{acc}(B_m) - \mathrm{conf}(B_m)\,\bigr|.
\end{equation}
A perfectly calibrated model has $\mathrm{ECE}=0$.

\paragraph{Brier score.}
The mean squared error between predicted probabilities and outcomes,
$\mathrm{Brier} = \frac{1}{N}\sum_{i}\sum_{c}(\hat{p}_{ic} - y_{ic})^2$, summing
over classes $c$; lower is better.

\paragraph{Neighborhood purity@$k$.}
Used in the representation analysis. For a query visit $i$ with $k$ nearest
neighbors $\mathcal{N}_k(i)$ under cosine distance in the representation, and a
categorical label $y$,
\begin{equation}
\mathrm{purity}@k = \frac{1}{N}\sum_{i} \frac{1}{k}\sum_{j\in\mathcal{N}_k(i)} \mathbf{1}[\,y_j = y_i\,],
\end{equation}
the mean fraction of a visit's neighbors sharing its label.

\section{Baseline Descriptions}
\label{app:baselines}

\paragraph{Sequential clinical models.}
\begin{itemize}
\item \textbf{GRU}~\cite{medsker2001RNN}: a gated recurrent network encoding a patient's visit sequence.
\item \textbf{Transformer}~\cite{vaswani2017transformer}: self-attention over the visit sequence.
\item \textbf{Deepr}~\cite{nguyen2016deepr}: a convolutional reader over a linearized sequence of medical codes.
\item \textbf{ConCare}~\cite{ma2020concare}: a time-aware recurrent model with feature-level self-attention that captures the personal clinical context.
\item \textbf{Dr.~Agent}~\cite{gao2020dragent}: a reinforcement-learning predictor with two policy agents that mimic clinical second opinions.
\item \textbf{AdaCare}~\cite{ma2020adacare}: a dilated-convolutional model with scale-adaptive feature recalibration capturing biomarker dynamics.
\item \textbf{StageNet}~\cite{gao2020stagenet}: a stage-aware LSTM with a convolutional module modeling disease-progression stages.
\item \textbf{GRASP}~\cite{zhang2021grasp}: a cohort-based model that leverages clinically similar patients for health-status representation.
\end{itemize}

\paragraph{Graph- and ontology-based EHR models.}
\begin{itemize}
\item \textbf{GRAM}~\cite{choi2017gram}: attention over the ICD diagnosis ontology so that rare codes borrow representation from their ancestors.
\item \textbf{GraphCare}~\cite{jiang2023graphcare}: personalized patient knowledge graphs drawn from external and language-model knowledge bases.
\item \textbf{MulT-EHR}~\cite{chan_multitask_ehr}: multi-task learning on a heterogeneous EHR graph with a causal denoising module; our reference baseline.
\end{itemize}

\paragraph{Medication-specific models.}
\begin{itemize}
\item \textbf{MICRON}~\cite{yang2021micron}: a recurrent residual network predicting medication changes between visits.
\item \textbf{MoleRec}~\cite{yang2023molerec}: molecular substructure-aware representation learning for drug recommendation.
\item \textbf{SafeDrug}~\cite{yang2021safedrug}: dual molecular graph encoders recommending effective, interaction-safe drug combinations.
\end{itemize}

\section{Hyperparameters and Training Details}
\label{app:hyperparams}

We report the settings of all three stages of the method (graph construction,
feature pretraining, and encoder training). Values are shared across both
databases unless noted otherwise.

\paragraph{Graph construction.}
Co-occurrence edges are retained under the co-occurrence criterion, based on
normalized pointwise mutual information (NPMI), with NPMI threshold $\tau=0.10$
and count floor $\kappa=5$.

\paragraph{Feature pretraining.}
Clinical-concept features are Bio\_ClinicalBERT embeddings, mean-pooled over
tokens and reduced to $128$ dimensions by PCA; the language model is frozen.
Patient and visit features are TransE embeddings of dimension $128$, trained on
the membership and temporal relations with margin $0.1$, learning rate
$5\times10^{-3}$, and $5$ uniformly-sampled negatives per edge; only patient and
visit embeddings are updated, and all node features are then held fixed.

\paragraph{Encoder and optimization.}
The encoder uses $L=2$ message-passing layers, hidden width $d=128$ (input
dimension $d_{\mathrm{in}}=128$), and $H=8$ attention heads, with feature dropout
$0.3$ and no input dropout. Training uses Adam with weight decay $10^{-5}$ for
$1000$ epochs on a $90\%/10\%$ visit-level split, seed $612$. Tasks are balanced
with DB-MTL gradient balancing using momentum
$\rho=0.9$ and $\varepsilon=10^{-8}$. The drug-recommendation logits are
temperature-scaled during training, with the temperature annealed to a floor of
$0.1$; the remaining tasks use their raw logits. Cross-visit temporal attention
aggregates over each visit's full set of preceding admissions. Two settings
differ by database, reflecting their sizes:
\begin{center}
\small
\begin{tabular}{lcc}
\toprule
 & MIMIC-III & MIMIC-IV \\
\midrule
Learning rate & $3\times10^{-4}$ & $5\times10^{-4}$ \\
Batch size    & $4{,}096$        & $4{,}096$ \\
\bottomrule
\end{tabular}
\end{center}

\paragraph{Checkpoint selection.}
The reported checkpoint is the epoch maximizing the mean of the four test
AUROCs; every task is read from that same checkpoint.

\paragraph{Temperature Annealing.}
The multi-label drug-recommendation loss vanishes early in training as the
sigmoid saturates. We alleviate this by scaling the drug logits with a
temperature annealed over epochs to a floor of $0.1$, which keeps the loss
informative without altering the task ranking.

\paragraph{Subgraph Sampling.}
The full EHR graph does not fit in memory, especially for MIMIC-IV. At each step
we therefore sample a batch of visit nodes together with their connected
neighbors, forming a subgraph over which messages are passed; we use $4{,}096$
visits per batch on MIMIC-III and MIMIC-IV.

\paragraph{Downsampling for Imbalanced Tasks.}
The binary tasks are heavily imbalanced, most severely mortality, where survivors
vastly outnumber deaths. During training we downsample the majority class to
match the minority, balancing the batches for these tasks.

\section{MIMIC-IV Post-hoc Analysis}
\label{app:posthoc_mimic4}

The post-hoc analysis in the main paper uses the frozen MIMIC-III model as a
representative case. Here we repeat the same analysis on the frozen MIMIC-IV
model, our canonical model. As before, we do not retrain the model. We read the
visit representation at four stages, $\mathrm{L}_{in}$ (pretrained input),
$L_0$ (after input projection and temporal encoding), and $L_1,L_2$ (after the
two message-passing layers), and evaluate on the held-out MIMIC-IV test split.
The findings agree with MIMIC-III on two of the three properties and reveal one
honest limitation on the third (mortality calibration).

\paragraph{Task signal becomes linearly accessible with depth.}
We fit a linear probe to each stage and read off how well each task is linearly
recoverable (Table~\ref{tab:probe_m4}). For every task the probe improves from
the pretrained input to the final layer, so the message-passing layers make the
representation more informative for all four tasks at once, matching the trend on
MIMIC-III.

\begin{table}[t]
\centering
\small
\setlength{\tabcolsep}{5pt}
\caption{Linear-probe performance on MIMIC-IV by stage. Values are AUROC, except
length of stay (macro one-versus-one AUROC), drug recommendation
(sample-averaged AUROC), and ICD chapter (accuracy over $19$ chapters). Every
task improves from the pretrained input ($\mathrm{L}_{in}$) to the final graph
layer ($L_2$).}
\label{tab:probe_m4}
\begin{tabular}{lcccc}
\toprule
Target & $\mathrm{L}_{in}$ & $L_0$ & $L_1$ & $L_2$ \\
\midrule
Mortality (AUROC)        & 0.625 & 0.627 & 0.734 & \textbf{0.747} \\
Readmission (AUROC)      & 0.704 & 0.718 & 0.754 & \textbf{0.762} \\
Length of stay (OvO)     & 0.673 & 0.656 & 0.773 & \textbf{0.800} \\
Drug rec.\ (sample)      & 0.964 & 0.961 & 0.971 & \textbf{0.973} \\
\midrule
ICD chapter (accuracy)   & 0.447 & 0.386 & 0.490 & \textbf{0.531} \\
\bottomrule
\end{tabular}
\end{table}

\paragraph{Neighborhoods are organized by outcome.}
For each stage we measure how often a visit's $10$ nearest neighbors share its
label (Table~\ref{tab:knn_m4}). Readmission neighborhoods grow steadily purer
with depth ($0.537\to0.639$), and the two clinical-structure controls, ICD
chapter and comorbidity burden, also rise. Mortality purity is high at every
stage and barely changes, because deaths are rare on MIMIC-IV (about $1.6\%$ of
test visits), so a neighborhood is almost always dominated by survivors and the
number is driven by the base rate rather than by depth.

\begin{table}[t]
\centering
\small
\setlength{\tabcolsep}{5pt}
\caption{Neighborhood purity at $k{=}10$ on MIMIC-IV by stage: the mean fraction
of a visit's $10$ nearest neighbors that share its label. Readmission and the
clinical-structure controls increase with depth; mortality is near-constant and
reflects the low death rate rather than representation quality.}
\label{tab:knn_m4}
\begin{tabular}{lcccc}
\toprule
Label & $\mathrm{L}_{in}$ & $L_0$ & $L_1$ & $L_2$ \\
\midrule
Readmission          & 0.537 & 0.606 & 0.639 & \textbf{0.639} \\
ICD chapter          & 0.276 & 0.174 & 0.316 & \textbf{0.375} \\
Comorbidity burden   & 0.323 & 0.284 & 0.452 & 0.437 \\
Mortality            & 0.966 & 0.967 & 0.969 & 0.969 \\
\bottomrule
\end{tabular}
\end{table}

\paragraph{Clinical concepts are linear directions.}
We fit a linear probe at $L_2$ to the presence of each body-system concept group,
where each MIMIC-IV diagnosis (a CCS category) is assigned to an ICD chapter by
the crosswalk of Appendix~\ref{app:vocab}. Concepts are strongly linearly
recoverable (Table~\ref{tab:cav_m4}): most groups exceed $0.75$ AUROC, led by
pregnancy and childbirth ($0.994$), and even the most frequent group, the
circulatory system ($86{,}201$ visits), reaches $0.892$. Recoverability is
higher and more uniform than on MIMIC-III, consistent with the far larger MIMIC-IV
cohort giving each concept more support.

\begin{table}[t]
\centering
\small
\setlength{\tabcolsep}{5pt}
\caption{Concept-probe AUROC at $L_2$ on MIMIC-IV. $n$ is the number of visits
with at least one diagnosis in the group. Concepts are recovered as linear
directions in the representation.}
\label{tab:cav_m4}
\begin{tabular}{@{}lrc@{}}
\toprule
Concept group & $n$ & Probe AUROC \\
\midrule
Pregnancy \& childbirth        & 19{,}988 & \textbf{0.994} \\
Digestive system              & 25{,}193 & 0.898 \\
Neoplasms                     & 15{,}498 & 0.897 \\
Musculoskeletal \& connective & 12{,}321 & 0.894 \\
Circulatory system            & 86{,}201 & 0.892 \\
Respiratory system            &  7{,}837 & 0.878 \\
Genitourinary system          &  4{,}957 & 0.876 \\
Injury \& poisoning           & 14{,}523 & 0.863 \\
Skin \& subcutaneous tissue   &  2{,}019 & 0.837 \\
External causes (E)           &  5{,}942 & 0.825 \\
Nervous system \& sense organs&  7{,}258 & 0.816 \\
Infectious \& parasitic       &  4{,}744 & 0.799 \\
Mental disorders              &  8{,}936 & 0.776 \\
Endocrine / metabolic         & 30{,}814 & 0.760 \\
Symptoms \& ill-defined       &  6{,}762 & 0.753 \\
Blood \& blood-forming        &  2{,}306 & 0.747 \\
Supplementary (V)             & 21{,}787 & 0.733 \\
Congenital anomalies          &     351  & 0.577 \\
\bottomrule
\end{tabular}
\end{table}

\paragraph{Calibration is mixed on MIMIC-IV.}
Table~\ref{tab:calib_m4} reports the expected calibration error (ECE) of each
task at $L_2$. Length of stay ($0.018$) and drug recommendation ($0.003$) are well calibrated. Readmission is
overconfident in its raw probabilities ($0.241$).
Mortality is the exception: its raw ECE is high ($0.461$). Its ranking remains useful (AUROC $0.788$), but its probabilities should
not be read as calibrated risks without a prevalence-aware recalibration. We flag this high calibration error for mortality prediction as a limitation.

\begin{table}[t]
\centering
\small
\setlength{\tabcolsep}{6pt}
\caption{Calibration on the MIMIC-IV test split at $L_2$. ECE is the expected
calibration error (lower is better).}
\label{tab:calib_m4}
\begin{tabular}{lcc}
\toprule
Task & AUROC & ECE (raw) \\
\midrule
Length of stay        & 0.792 & 0.018  \\
Drug recommendation   & 0.968 & 0.003 \\
Readmission           & 0.970 & 0.241  \\
Mortality             & 0.788 & 0.461 \\
\bottomrule
\end{tabular}
\end{table}

\bibliography{aaai2027}

\end{document}